\documentclass[letterpaper, 10 pt, conference]{ieeeconf}  

\IEEEoverridecommandlockouts    

\usepackage{times}

\usepackage{amsmath}
\usepackage{amssymb}
\usepackage{amsthm}
\usepackage[ruled, vlined, linesnumbered]{algorithm2e}
\usepackage{graphicx}
\usepackage[dvipsnames]{xcolor}
\usepackage{adjustbox}
\usepackage{array}
\usepackage{multirow}
\usepackage{multicol}
\usepackage{caption}

\usepackage{subcaption}
\usepackage[bookmarks=true]{hyperref}
\usepackage{cleveref}
\usepackage{paralist}
\usepackage{tikz}
\usetikzlibrary{automata, positioning, arrows}
\tikzset{initial text={}} 
\usepackage{cite}
\usepackage{booktabs}
\usepackage{tabularx}
\usepackage{multirow}
\usepackage{xspace}
\usepackage{adjustbox}

\usepackage{eso-pic}

\newtheorem{problem}{Problem}
\newtheorem{myexam}{Example}

\newtheorem{definition}{Definition}
\newtheorem{remark}{Remark}

\newcommand{\px}{\mathbf{x}}
\newcommand{\z}{\mathbf{z}}

\newcommand{\prop}{\mathrm{p}}
\newcommand{\Aphi}{\mathcal{A}_{\varphi}}
\newcommand{\Aphis}{\mathcal{A}_{\neg\varphi_S}}

\newcommand{\str}{\pi}

\newcommand{\p}{\mathcal{P}}
\newcommand{\U}{\mathcal{U}}
\newcommand{\X}{\mathcal{X}}
\newcommand{\F}{\mathcal{F}}
\newcommand{\G}{\mathcal{G}}

\newcommand{\reals}{\mathbb{R}}
\newcommand{\nats}{\mathbb{N}}

\DeclareMathOperator*{\argmin}{\arg \min}
\DeclareMathOperator*{\argmax}{\arg \max}

\newcommand{\rr}[1]{\textcolor{red}{[RR: #1]}}

\title{\LARGE \bf Shielded Deep Reinforcement Learning for Complex Spacecraft Tasking}

\author{Robert Reed, Hanspeter Schaub, and Morteza Lahijanian
\thanks{This work was supported by Air Force Research Lab (AFRL) under agreement number FA9453-22-2-0050. 
}
\thanks{Authors are with the Department of Aerospace Engineering Sciences at University of Colorado Boulder, Boulder, Colorado, USA. {\tt\small \{Robert.Reed-1, Hanspeter.Schaub, Morteza.Lahijanian\}@colorado.edu }}%
}

\begin{document}

\AddToShipoutPictureBG*{%
  \AtPageUpperLeft{%
    \hspace{15.1cm}%
    \raisebox{-1.5cm}{%
      \makebox[0pt][r]{To appear in the American Control Conference, July 2024.}}}}

\maketitle
\thispagestyle{empty}
\pagestyle{empty}

\begin{abstract}
Autonomous spacecraft control via Shielded Deep Reinforcement Learning (SDRL) has become a rapidly growing research area. However, the construction of shields and the definition of tasking remains informal, resulting in policies with no guarantees on safety and ambiguous goals for the RL agent. In this paper, we first explore the use of formal languages, namely Linear Temporal Logic (LTL), to formalize spacecraft tasks and safety requirements.  We then define a manner in which to construct a reward function from a co-safe LTL specification \emph{automatically} for effective training in SDRL framework.
We also investigate methods for constructing a shield from a safe LTL specification for spacecraft applications and propose three designs that provide probabilistic guarantees.  We show how these shields interact with different policies and the flexibility of the reward structure through several experiments.
\end{abstract}


\section{Introduction}
    \label{sec:intro}

The spacecraft task scheduling problem, which involves collecting data while adhering to system constraints, traditionally relies heavily on human intervention. This reliance is due to the inability of current autonomy modules, often based on simplistic rules and past experiences, to ensure compliance with spacecraft safety requirements. 
Due to recent technological advancements and economical interest~\cite{frost2010challenges}, spacecraft autonomy has become a central research topic \cite{harris2020spacecraft, nazmy2022shielded, adams2023overview}, and 
recent works show that computational challenges such as the large dimensionality of the state space and low on-board computation capabilities can be overcome via Reinforcement Learning (RL) \cite{nazmy2022shielded, harris2020spacecraft}. 
Nevertheless, providing correctness and safety guarantees on the decisions of the autonomy remains a major challenge.  This work focuses on this challenge and aims to enable safe spacecraft autonomy by combining machine learning with formal methods.

In RL, an agent explores an unknown environment and acts to maximize a reward function that is designed to express the desired behavior of the agent. Typically, this reward function is hand designed. Deep RL (DRL) is an extension of RL that utilizes the power of Neural Networks (NNs) to learn a policy, enabling RL in high-dimensional spaces. While the optimally and data-efficiency of DRL algorithms is well understood \cite{arulkumaran2017deep}, the policies returned from these algorithms have no guarantees on safety. 

This led research in the direction of Shielded DRL (SDRL) \cite{bloem2015shield, alshiekh2018safe} with adaptation for spacecraft autonomy \cite{harris2020spacecraft,nazmy2022shielded}.
In SDRL, a \emph{shield} is designed in order to ensure \emph{safety} of the system when it is deployed with a DRL policy \cite{alshiekh2018safe}. The shield acts as a \emph{minimal interference} filter on the agents action, allowing all actions that have been pre-determined to be safe and replacing unsafe actions with correct choices. SDRL had been shown to improve policy performance and reduce the necessary training time for spacecraft operations \cite{nazmy2022shielded, harris2020spacecraft}.
Typically, shield design relies on having a description of the safety-critical aspects of the system as a Markov Decision Process (MDP), which is called a \emph{Safety MDP}. Prior works~\cite{nazmy2022shielded} use a hand-designed safety MDP, where the transition probabilities between states are chosen from expert intuition.  Such a design not only requires a domain expert with extensive knowledge, but also limits the guarantees on safety that the shield can provide.

Formal languages, such as \emph{linear temporal logic} (LTL)~\cite{pnueli1977temporal}, provide a manner to rigorously define the tasks and safety requirements needed for spacecraft deployment. Typically, the tasks that an agent should complete are composed as a \emph{liveness specification} and the behaviors that should be avoided are written as a \emph{safety specification}. Formal specifications have been used with great success in robotics applications \cite{bhatia2010sampling, bhatia2011motion} and the effectiveness of splitting the specifications into parts (i.e., liveness and safety) has been well demonstrated \cite{lahijanian2016iterative}. This separation follows the idea of constructing a shield based on a safety specification and learning a policy that satisfies a liveness specification. When a task is defined using LTL, formal synthesis techniques \cite{kress2018synthesis, lahijanian2015formal} are often used to find a policy that is correct-by-construction. However, such techniques are often limited by the dimensionality of the system, hence the need for DRL in the autonomous spacecraft scenario.

This work focuses on incorporating formal methods into the SDRL framework for the autonomous Earth imaging problem by formalizing the design of a shield for a spacecraft system and utilizing LTL formulas to define tasking and safety requirements. We propose to construct a safety MDP algorithmically using physics-simulator engines and implement three different shield designs with this MDP. Prior works in formal methods often assume the safety MDP is known a priori~\cite{bloem2015shield}, or restrict the RL to a set of safe policies and work to optimize an unknown objective~\cite{wen2015correct, junges2016safety}, unlike our scenario. We also identify a manner to automatically construct a reward function from an LTL specification, removing human interpretation of a task from the process. This ensures that DRL is optimizing a reward function that has no ambiguity from the desired task, resulting in a policy that is optimized for the desired task.  Our evaluations show the effectiveness of the reward and shield designs and the importance of training with safety specifications.

Our contributions are four-fold: 1) we improve the formalism of shield construction for spacecraft SDRL, 2) we demonstrate how to incorporate complex, formal specifications for Earth imaging tasks into a DRL framework, 3) we identify a training setup that minimizes safety violations with few shield interventions, and 4) we illustrate the efficacy of the method on several case studies and benchmarks.

\section{Problem Formulation}
    \label{sec:problem}
    In this work, we consider autonomous spacecraft scheduling for complex Earth observation tasks with a discrete action space.
The spacecraft must select a sequence of flight modes such that a predefined Earth observing task is satisfied while remaining safe. 

\subsection{Spacecraft Model}

Spacecraft dynamics are highly complex and high dimensional, with potentially thousands of states needed to accurately represent how each of the subsystems on a spacecraft interact.  The dynamics act over continuous space and time with stochastic disturbances (e.g. external torques), which enhance the difficulty of control problems.
We assume that we are able to control the switching between different modes of operation. 
Hence, the dynamics of the spacecraft system can be described as a continuous-state Markov decision process (MDP).

\begin{definition} [MDP] \label{def:mdp}
    A \emph{Markov Decision Process} (MDP) is a tuple $M = (X, x_0, A, T, \Pi, L)$, where
    \begin{itemize}
        \item $X \subseteq \reals^n$ is the state space,
        \item $x_0 \subset X$ is a set of initial states,
        \item $A$ is a finite set of modes or actions,
        \item $T: X \times A \times \mathcal{B}(X) \rightarrow [0, 1]$ is a transition probability function, where $\mathcal{B}$ is a Borel set\footnote{The Borel set defines an open set of states in $X$, hence transition probabilities can be assigned.},
        \item $\Pi$ is a set of atomic propositions that are related to spacecraft task or safety, and
        \item $L: X \rightarrow 2^\Pi$ is a labeling function that assigns a state $x \in X$ to a subset of $\Pi$.
    \end{itemize} 
\end{definition}

\begin{myexam}\label{ex:sat}
    One can consider a spacecraft with fours modes of operation $A = \{a_i\}_{i=0}^3$, where
    $a_0$ is \emph{Charging Mode}, $a_1$ is \emph{Momentum Dumping Mode}, $a_2$ is \emph{Imaging Mode A}, and $a_3$ is \emph{Imaging Mode B}. 
    The Earth observation tasking can be related to the two imaging modes. The Momentum Dumping and Charging modes are then important from the perspective of safety.
\end{myexam}

Defining the state of the system as $x \in X$, an infinite trajectory is then written as $\omega_\px = x_0 \xrightarrow{u_0} x_1 \xrightarrow{u_1} ...$ where each $u_i \in A$.  We denote the $i$-th element of $\omega_\px$ by $\omega_\px[i]$, and the set of all finite and infinite trajectories are $\Omega_\px^\text{fin}$ and $\Omega_\px$, respectively. We are interested in controlling the trajectory through the choice of action taken (switching modes) via a policy.

\begin{definition} [Policy]
    A \textit{policy} $\str: \Omega_\px^\text{fin} \rightarrow A$ is a function that maps a finite trajectory $\omega^N_\px \in \Omega_\px^\text{fin}$ onto the next action in $A$.  Policy $\pi$ is called stationary if it only depends on the last element of 
    $\omega_\px^N$; otherwise, it is called history dependent.
\end{definition}

Under a policy $\str$, a probability measure over the paths of the MDP $M$ is well defined \cite{lahijanian2011control}. We denote MDP $M$ under $\str$ as $M^\str$ and its sets of finite and infinite trajectories as $\Omega_\px^{\text{fin},\pi}$ and $\Omega_\px^{\pi}$, respectively.

\subsection{LTL for Earth Observing Tasks and Safety Requirements}

We consider imaging tasks that can be completed in finite time and safety requirements that must not be violated. 
These tasks and requirements are related to the temporal behavior of the spacecraft system with respect to a set of state-space regions $R = \{r_1, \ldots, r_l\}$, where $r_i \subseteq X$. 
To enable formal description of tasks,
we associate an atomic proposition $\prop_i$ to each region $r_i$ such that $\prop_i$ is true iff $x \in r_i$.  Then, the set of atomic proposition is $\Pi = \{\prop_1, \ldots, \prop_l\}$, and the labeling function $L: X \to 2^\Pi$ assigns each state $x \in X$ to the set of atomic propositions that are true at that state. Accordingly, we define an \textit{observation trace} of trajectory $\omega_\px$ to be $\rho = \rho_0 \rho_1 \ldots$, where $\rho_i = L(\omega_\px[i])$ for all $i \geq 0$.

To formally specify spacecraft requirements, we use \textit{co-safe} and \textit{safe} LTL~\cite{kupferman2001model}, which are languages that can express the temporal behaviors of a system with a set of Boolean connectives and temporal operators. Co-safe LTL is used to describe tasks that the spacecraft should achieve.

\begin{definition}[Co-safe LTL]
    \label{def:csLTL}
    Given a set of atomic propositions $\Pi$, a co-safe LTL formula is recursively defined as
    \begin{equation*}
        \varphi = \prop \mid \neg \prop \mid \varphi \land \varphi \mid \X \varphi \mid  \varphi \U \varphi  \mid \F \varphi
    \end{equation*}
    where $\prop \in \Pi$,  $\neg$ (``not'') and $\land$ (``and'') are Boolean connectives, and  $\X$ (``next''), $\U$ (``until''), and $\F$ (``eventually'') are temporal operators.
\end{definition}

Safe LTL are then used to define behaviors that the spacecraft should avoid.
\begin{definition}[Safe LTL]
    \label{def:sLTL}
    Given a set of atomic propositions $\Pi$, a safe LTL formula is inductively defined as
    \begin{equation*}
        \varphi = \prop \mid \neg \prop \mid \varphi \land \varphi \mid \X \varphi \mid \G \varphi 
    \end{equation*}
    where $\prop \in \Pi$,  $\neg$, $\land$, and $\X$ are as in Definition \ref{def:csLTL} and $\G$ (``globally'') is a temporal operator.
\end{definition}

The semantics of safe and co-safe LTL are defined over infinite traces \cite{kupferman2001model}.  We say an infinite trajectory $\omega_\px \in \Omega_\px$ satisfies an LTL formula $\varphi$, denoted as $\omega_\px \models \varphi$, if its trace satisfies $\varphi$.
For our problem, we consider specifications of form 
$$\varphi = \varphi_L \land \varphi_S,$$ 
where $\varphi_L$ is a \textit{liveness} specification that describes the task that the spacecraft should achieve given as a co-safe LTL formula, and $\varphi_S$ is a \textit{safety} specification that identifies what the spacecraft must avoid as a safe LTL formula. While the satisfaction of co-safe LTL formula are defined over infinite trajectories, we can assess if a co-safe LTL formula is satisfied with a finite trajectory. Similarly, we can only assess satisfaction of a safe LTL formula over infinite trajectories, but the negation of a safe LTL formula ($\lnot \varphi_S$) is a co-safe LTL formula, whose satisfaction can be assessed on finite trajectories. Then, if a finite trajectory satisfies $\lnot \varphi_S$, it cannot satisfy $\varphi_S$, i.e., the trajectory violates $\varphi_S$. Hence,
we say a length $N \in \nats$ prefix of $\omega_\px$, denoted as $\omega_\px^N$, satisfies formula $\varphi = \varphi_L \land \varphi_S$, denoted by $\omega_\px^N \models \varphi$ iff $\omega_\px^N$ does not violate $\varphi_S$
and satisfies $\varphi_L$, i.e., 
$$\omega^N_\px \models \varphi_L \land \varphi_S  \quad \text{iff} \quad \omega^N_\px \not\models \neg \varphi_S \land \exists i\leq N, \, \omega^i_\px \models \varphi_L.$$
The combination of safe and co-safe LTL enables the description of a wide variety of tasks.

Recall that the trajectories of $M$ under $\pi$ have an associated probability measure; hence, satisfaction of $\varphi$ is probabilistic. The probability of trajectories of $M$ satisfying formula $\varphi$ under $\str$ is defined as $$P(M^\str \models \varphi) = P(\omega_\px \in \Omega_\px^{\text{fin},\str} \mid \omega_\px \models \varphi).$$


\begin{myexam} \label{ex:spec}
    Consider the spacecraft in Example~\ref{ex:sat} with the  goal of imaging a target location on Earth with constraints on how the dynamics affect the image quality. In natural language, this task may be written as ``Image the target successfully, and only accept images that are taken when the attitude error ($|{\sigma}_{err}|$) is less than 0.008 radians and the attitude rate ($|\dot{\sigma}|$) is less than 0.002 radians per second". With a signal $i \in \{0, 1\}$ that identifies if the target is accessible for imaging, we then define the region $r_0$ as a subset of state space $X$ as 
    $r_0= \{x\in X \mid 0 \leq |{\sigma}_{err}| < 0.008, \hspace{1mm} 0 \leq |\dot{\sigma}| < 0.002, \hspace{1mm} a \in \{a_2, a_3\}, \hspace{1mm} i = 1\}$.
    We associate the atomic proposition $\prop_0$ with region $r_0$ and say $\prop_0$ is true iff the state of the spacecraft $x \in r_0$; otherwise $\prop_0$ is false. Our task can then be written in co-safe LTL as 
    \begin{equation}
        \label{eq:simple_liveness_spec}
        \varphi_{0L} = \F \prop_0,
    \end{equation}
    meaning eventually have the state of the spacecraft be in region $r_0$. A finite trajectory of the spacecraft $\omega_\px^N$ can be automatically assessed to identify if the specification is satisfied by checking the observations associated with each state, i.e. $\rho_i = L(\omega_\px^N(i))$. If $\prop_0 \in \rho_i$ for some $i \leq N$, then the trajectory satisfies the specification, written as $\omega_\px^N \models \varphi_{0L}$.

    Consider the addition of a safety requirement to the specification as ``Additionally, never allow power to fall below 20\% and never allow reaction wheel speeds above 80\% of their maximum." The safety specification is concerned with stored charge $\p$ and reaction wheel speeds $\Omega$. We define regions $r_1: \{x\in X \mid 0 \leq \p < 0.2\}$ and $r_2: \{x\in X \mid 0.8 < \Omega \leq 1\}$ and corresponding atomic propositions $\prop_1, \prop_2$, resulting in a safe LTL specification
    \begin{equation}
        \label{eq:simple_safety_spec}
        \varphi_S = \G(\lnot(\prop_1 \vee \prop_2)).
    \end{equation}
    The task would then only be considered satisfied if a finite trajectory $\omega_\px^N \models \varphi_L \land \varphi_S$.
    More complex specifications can be defined in a similar manner.
\end{myexam}


\subsection{Problem Statement}
Our problem is then be defined as follows.
\begin{problem} [Safe Control]
Given a model of the system as an MDP $M$, an Earth observation task specified as a co-safe LTL formula $\varphi_{L}$, a safety specification as a safe LTL formula $\varphi_S$, and (safety) probability threshold $p$,
find a policy $\str^*$ such that the probability of satisfying $\varphi_L \land \varphi_S$ is maximized while the probability of violating $\varphi_S$ is less than or equal to $p$, i.e.,
$$\str^* = \argmax_\str P(M^{\str} \models \varphi_L \land \varphi_S)$$ 
subject to
$$P(M^{\str^*} \models \neg \varphi_S) \leq p.$$ 


\end{problem}

Note that there are several challenges in the above problem.  Firstly, the state space for a spacecraft can contain thousands of state parameters, resulting in a dimensionality that is too large for traditional synthesis techniques. To overcome the problem of \emph{state explosion}, we use DRL, with the policy being captured as a NN which enables generalizability.  
This is possible because a high-fidelity, flight-approved spacecraft physics simulator exists, namely the Basilisk\footnote{https://hanspeterschaub.info/basilisk} astrodynamics simulation framework~\cite{kenneally2020basilisk}, which enables DRL for spacecraft with its computational performance and flexible framework for integration with common machine learning techniques~\cite{harris2020spacecraft}.  

However, using a NN policy that is trained via DRL introduces a new challenge: 
how can the satisfaction of the safety constraint during spacecraft operation be ensured?
To address this challenge, we employ shielded DRL (SDRL) in which a shield is constructed from $\varphi_S$ and deployed with the policy to ensure the correctness of actions taken by the spacecraft. 
Finally, shield design itself for a spacecraft is a challenge due to large dimensionality.  This often leads to overly conservative-shields which can interrupt task execution unnecessarily and too frequently, significantly reducing efficiency.  We propose several methods of shield design and training that mitigate this problem.

\begin{remark}
   While our approach is general, in this paper, we follow the observation space setup from \cite{harris2020spacecraft} which allows the use of DRL for the spacecraft-planning problem, i.e., the observation space consists of canonical position and velocity vectors, attitude error, attitude rate, wheel speeds, stored charge, a sun access indicator, and a target access indicator. 
\end{remark}

\section{Reward and Shield Design for DRL}
    \label{sec:shield_design}
    
\begin{figure}[t]
    \centering
    \includegraphics[width=0.8\linewidth]{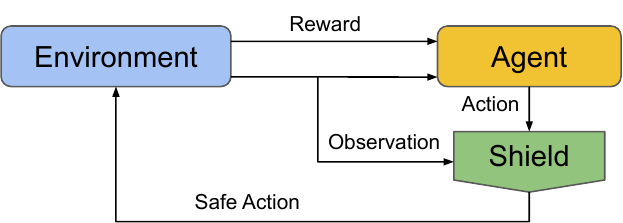}
    \caption{Post-Posed Shielded RL architecture.}
    \label{fig:shield_arch}
\end{figure}

In this section, we discuss how to perform DRL with LTL specifications and describe how we construct shields for the spacecraft system. 
In Post-Posed SDRL, as in Figure~\ref{fig:shield_arch}, a shield acts as a monitor on the actions of the learning agent. The shield allows any action that is safe and corrects actions that are unsafe before returning the choice to the environment. The shield is referred to as post-posed due to the ordering of acting after the learning agent. 

\subsection{Rewards for DRL with LTL Specifications}
The goal of RL is to find an optimal, or nearly-optimal, policy that maximizes the expected value of a reward function. That is, given an MDP $M$ and a reward function $R: X \times A \to \reals$ find a policy $\str^* = \argmax_{\str} \mathbb{E}[\sum R(s, a)]$. 
Traditionally, the reward function is manually constructed to define the task that the agent is learning. The manual construction of a reward function is prone to error, resulting in a disconnect between what the agent learns and the desired results. To address this issue, we write the objective in (co-safe) LTL and automatically construct a reward function from the specification. Work \cite{hahn2023mungojerrie} defines a method to construct a reward function from an LTL specification, and provides a relation between the probability of satisfying the specification with the expected value of the reward. 
To use this method, a deterministic finite automaton (DFA) must be constructed from the co-safe LTL specification, i.e., $\varphi_L$ or $\neg \varphi_S$,
that accepts the same traces as $\varphi$ \cite{kupferman2001model}. 

\begin{definition} [DFA]  \label{def:dfa} 
A deterministic finite automaton (DFA) constructed from an LTL formula $\varphi$ is a tuple $\Aphi = (Z, z_0, 2^{\Pi}, \delta, Z_f)$, where 
\begin{itemize}
    \item $Z$ is a finite set of states,
    \item $z_0 \in Z$ is an initial state,
    \item $2^{\Pi}$ is a finite set of input symbols,
    \item $\delta : Z \times 2^{\Pi} \rightarrow Z$ is a transition function, and
    \item $Z_f \subseteq Z$ is the set of final (accepting) states.
\end{itemize}

\end{definition}

A finite run on $\Aphi$ is a sequence of states $\z = z_0z_1...z_{n+1}$ induced by a trace $\rho = \rho_0\rho_1...\rho_n$ where $\rho_i \in 2^{\Pi}$ and $z_{i+1} = \delta(z_i, \rho_i)$. A finite run is accepting if, for some $i\leq n$, $z_i \in Z_f$. If a run is accepting, its associated trace is accepted by $\Aphi$. 
The set of all traces that are accepted by $\Aphi$ is called the language of $\Aphi$. The language of $\Aphi$ is equal to the language of $\varphi$, i.e., trace $\rho$ is accepted by $\Aphi$ iff $\rho \models \varphi$.

To perform DRL given $\Aphi$ and MDP $M$, the product $M_{\Aphi} = M \times \Aphi$, which captures both the dynamics of $M$ and the constraints of $\varphi$, is necessary. 

\begin{definition}[Product MDP]
    \label{def:product MDP}
    The product MDP $M_{\Aphi} = M \times \Aphi$ is a tuple $M_{\Aphi} = (S, A, \Delta, S_f, S_v)$, where $S= X \times Z$ is the set of product states, $A$ is as in Def.~\ref{def:mdp}, $\Delta: S \times A \times (\mathcal{B}(X) \times Z) \rightarrow [0, 1]$ is the transition function such that $\Delta((x, z), a, (\mathcal{B}(x'), z') = T(x, a, \mathcal{B}(x'))$ if $z' = \delta(z, L(x))$ and is 0 otherwise, and $S_f = X \times Z_f$ 
    is the set of final states.
\end{definition}
With product construction in Def.~\ref{def:product MDP}, the states of $M_{\Aphi}$ encode the history of trajectories of $M$ with respect to $\varphi$, i.e., runs $\z$ on $\mathcal{\Aphi}$. Let $Z_v \subseteq Z\setminus Z_f$ be the set of DFA sink states, i.e., $\forall z \in Z_v$ and $\forall \sigma \in 2^\Pi$, $z = \delta(z,\sigma)$.
We adapt the method proposed in \cite{hahn2023mungojerrie} to our environment, and define the reward $R:S \to [-1,1]$, discount function $\Gamma \in (0,1)$, and cumulative reward function $V_F$ as:
\begin{align}
    &R((x,z), (x', z')) = 
    \begin{cases}
        1 - \gamma_F & \text{if $z' \in Z_f$} \\
        -1 & \text{if $z' \in Z_v$} \\
        1 - \gamma_T & \text{if $z \neq z'$} \\
        0 & \text{otherwise} \\
    \end{cases}, \label{eq:reward R} \\
    &\Gamma((x, z), (x', z')) = 
    \begin{cases}
        \gamma_F & \text{if $z' \in Z_f$} \\
        \gamma_T & \text{if $z \neq z'$} \\
        \gamma & \text{otherwise} \label{eq:reward gamma} \\
    \end{cases}, 
\end{align}
\begin{multline}
    V_F(\omega_\px, \z, n) = \sum_{i=0}^{n}R((\omega_\px[i], \z[i]), (\omega_\px[i+1], \z[i+1])) \\
    \prod_{j=0}^{i-1} \Gamma((\omega_\px[j], \z[j]), (\omega_\px[j+1], \z[j+1])),  \label{eq:reward V}
\end{multline}
where $\gamma, \gamma_T, \gamma_F \in (0, 1)$ are hyper-parameters, and $\z[i]$ is the $i$-th element of run $\z$. 
While this reward is defined over the product states, the product does not need to be constructed explicitly; it is sufficient to implicitly construct it for reward evaluation. 


Training with this reward is not conditioned on the presence of a shield and is then applicable for generic DRL tasks. As we train with co-safe LTL specifications which can be satisfied in finite time, it would be natural to terminate a training episode once an accepting state of the DFA is reached. In order to expand the search space, we instead reset to the initial state of the DFA if we reach an accepting state and allow the agent to continue training. Intuitively, this means a trajectory is more valuable if it can satisfy a specification quickly and frequently. This can result in rewards much larger than 1 in the event a specification can be satisfied multiple times in an episode. If we reach a sink state in $Z_v$ of the DFA, we cannot satisfy $\varphi$, and hence the episode is terminated and the reward is reduced by 1.

We note that the reward construction proposed in \cite{hahn2023mungojerrie} produces a direct relation between the probability of satisfying a specification and the cumulative reward received. However, reward is only received when reaching an accepting state of the DFA which can result in extremely sparse reward as the number of states in the DFA grows.
The alterations from the original design we propose in \eqref{eq:reward R}-\eqref{eq:reward V}, in particular, the additional reward in $R$ when $z \neq z'$, encourage transitions along the states of the DFA. However, they also limit the direct relation between reward and the probability of satisfying the specification. As noted in our case studies, a correlation between reward and satisfaction rate is retained.

For example, the DFA constructed from $\varphi_{0L}$ in Example~\ref{ex:spec} consists of only two states; an initial state and the accepting state. Here the reward formulation in \cite{hahn2023mungojerrie} performs identically to our modified version. However, if the specification is more complex, the resulting DFA is likely to have more states. In a specification that requires five images with alternating imaging modes and an inclusion of safety, the DFA has seven states. In our experiments, training to satisfy this specification with the original reward results in 38.9\% of trajectories ending with spacecraft failure, whereas our formulation results in only 1.4\% of trajectories failing; further discussion on these benchmarks can be seen in Section \ref{sec:experiments}. This difference is almost entirely due to the sparsity of reward, resulting in a sub-optimal policy when trained under the same number of epochs.

Lastly, we note that a policy $\pi: S \to A$ trained on product MDP $M_{\Aphi}$ is stationary.  This policy however becomes history dependent on $M$, i.e., since $S = X \times \Aphi$, the history is captured by the states of $\Aphi$.

\subsection{Shield Design}

Shield design is typically based on having an MDP that describes the evolution of the safety aspects of the system in question. In our problem, the MDP that describes the evolution of the spacecraft is unknown \emph{a priori}.  This poses a major challenge for designing a shield. 
To that end, we abstract $M$ to a finite MDP that fully captures the behavior of the spacecraft w.r.t. $\varphi_S$ and refer to it as the Safety MDP.  
We obtain this abstraction by partitioning state space $X$ such that the partition respects the regions of interest that correspond to $\varphi_S$.  Let $R_S \subseteq R$ be the set of safety regions.  Then, the Safety MDP is defined as follows.

\begin{definition} [Safety MDP] \label{def:finite_mdp}
    Given a partition of $X$ that respects the regions of interest in $R_S \subseteq R$,
    the \emph{safety} MDP is a finite-state MDP $\bar{M} = (Q, A, P, \bar{\Pi}, \bar{L})$, where $A$ is as in Def.~\ref{def:mdp}, 
    \begin{itemize}
        \item $Q = \{q_1, \ldots, q_m\}$ is a finite set of states obtained from the partition of $X$, i.e., $q_i \subseteq X$,
        \item $P: Q \times A \times Q \rightarrow [0, 1]$ is a transition probability function such that, for every $q,q' \in Q$ and $a \in A$, $P(q,a,q') = \mathbb{E}_{x\sim D(q)} [T(x,a,q')]$, where $D(q)$ is a probability distribution over region $q$,
        \item $\bar{\Pi} \subseteq \Pi$ is a set of (safety) atomic propositions that are associated with the regions in $R_S$,
        \item $\bar{L}: Q \rightarrow 2^{\bar{\Pi}}$ is a labeling function such that $\bar{L}(q) = L(x) \cap \bar{\Pi}$ for all $x \in q$.
    \end{itemize} 
\end{definition}

In general, the Safety MDP $\bar{M}$ does not require the full dimensionality of the MDP $M$. This dimensionality reduction enables rigorous safety analysis while maintaining computational tractability
when constructing a shield. 
Specifically, for the spacecraft in Example~\ref{ex:sat}, we are interested in regulating body rates $|\dot{\sigma}|$, reaction wheel speeds $\Omega$, and stored charge $\p$. As in RL, identifying an appropriate state space for the spacecraft is challenging, as the true state space may include thousands of states \cite{nazmy2022shielded} and the problem must remain Markovian~\cite{sutton2018reinforcement} in a lower dimensionality. 
We propose a reduction of the state space to just the values of interest: $|\dot{\sigma}|, \Omega, \p$. These states on their own are not Markovian. For instance, the attitude rate in the next time step cannot be predicted just from the current attitude rate as it also depends on states such as the attitude error. We approach this problem by identifying the transition probabilities between states of the safety MDP through simulation. 

We first define a domain $\bar{X} \subset X$ in which the spacecraft can safely operate, e.g., $|\dot{\sigma}| \leq 0.01$ , $\Omega \leq 1$ , and $0 < \mathcal{P} \leq 1$. We then discretize $\bar{X}$ into a set non-overlapping regions, defining the states $Q$ of $\bar{M}$. Due to the complexity of the dynamics, we compute the transition probabilities $P$ through simulation of $M$ using Basilisk~\cite{kenneally2020basilisk}. For each state $q \in Q$ of the MDP, we initialize the spacecraft simulation with randomized parameters (e.g. orientation) and limit the three states of interest to the values described by the discretization (e.g. $0 \leq \Omega < 0.2$) and simulate the evolution of the system $N_P$ times, where $N_P$ is a large number ($N_P = 10,000$ in our case studies) under each action. This allows the identified transition probabilities to no longer be conditioned on aspects such as orientation, allowing the abstraction of the safety MDP to remain Markovian.

\textbf{Shield Algorithms.}
With the safety MDP constructed above, shield synthesis is enabled. Given $\bar{M}$, the goal is to find the set of \emph{all} safe policies that guarantee no violation to $\varphi_S$ with at least probability $1-p$. As the policies can be history dependent, finding the set of all policies can lead to combinatorial problems and computational intractability.  Hence, we focus on finding a set of stationary policies on the product MDP with the DFA corresponding to $\varphi_S$.  

Since $\varphi_S$ is a safe LTL formula, $\neg \varphi_S$ is a co-safe formula.  Then, DFA $\Aphis$ can be constructed that precisely accepts all the safety-violating traces.  The product $\bar{M}_{\neg \varphi_S} = \bar{M} \times \Aphis$ can be constructed per Def.~\ref{def:product MDP} with the set of states $\bar{S}: Q \times Z$, transition probability function $\bar{\Delta}: \bar{S} \times A \times \bar{S} \to [0, 1]$, and set of final states $\bar{S}_f$. Note that the paths of $\bar{M}_{\neg \varphi_S}$ that reach $\bar{S}_f$ violate $\varphi_S$.  Hence, our aim becomes to find the set of stationary policies on $\bar{M}_{\neg \varphi_S}$, under which the probability of the paths that reach $\bar{S}_f$ is at most $p$. 

We note that the traditional shield construction methods as in \cite{bloem2015shield} are based on game formulation.
Those approaches strictly require all the paths of the MDP to remain safe all the time, which is analogous to requiring a violation probability threshold of $p=0$, which result in very conservative design.
For example, in our case studies, those approaches never produced a safe action. Therefore, we relax such strong requirement and focus on probabilistic shields, i.e., allowing a small violation probability up to a threshold $p$, and propose three different shield designs that result in safe actions for our MDP, each design providing a different guarantee on safety. 

\subsubsection{One-Step Safety}

The first shield we investigate is a one-time step safety shield, i.e., we find actions that 
enable transition to the safe set $\bar{S} \setminus \bar{S}_f$
with high probability. This shield is the simplest to implement, as actions that are safe on the product $\bar{M} \times \mathcal{A}_{\neg \varphi_{S1}}$ can be directly assessed from the transition probabilities of the MDP $\bar{M}$ given a state and action. This design provides a guarantee that the system remains safe with probability $1 - p$ for at least one time step; however, there are no guarantees on long term safety. We allow any action that transitions to a safe state, even if the next states have no safe actions themselves. Then the shield is defined as, for every $s \in \bar{S}$,
\begin{align*}
    \str_\textrm{\it shield}^1(s) = \{ a \in A \mid \sum_{s' \in \bar{S_f}}  \bar{\Delta}(s,a, s') < p \}.
\end{align*}

\subsubsection{Two-Step Safety}

As the first design has no long term safety guarantees, we design a shield that only allows actions that have a high probability of transitioning to safe states, where a safe state is recursively defined as a state where there is a safe action. This translates into finding actions that have a high probability of remaining safe for two time steps. This guarantees the system will remain safe for two consecutive time steps with probability $\geq1-p$ at each step. Two steps are considered sufficient as transition probabilities in MDPs are not history dependent; hence, this guarantee holds for each time step the system begins in a safe state. 
We call the set of unsafe states $U$ and initialize it with $U = \bar{S_f}$.  Then, states are recursively added to $U$ when no safe action is available. The process repeats until we reach a fixed point ($U$ gains no more states). The algorithm is as follows.
\begin{align*}
    &U = \bar{S}_f \\
    &\texttt{While } U \neq U'\\
    & \qquad U' = U \\
    & \qquad \forall s \in \bar{S},\; \str_\textrm{\it shield}^2(s) = \{ a \in A \mid \Sigma_{s' \in U}  \bar{\Delta}(s,a, s') < p \} \\
    & \qquad U = U \cup \{s \in \bar{S} \mid  \str_\textrm{\it shield}^2(s) = \emptyset\}\\
    &\texttt{Return } \bar{S} \setminus U, \; \str_\textrm{\it shield}^2
\end{align*}
In the event that the system transitions into the unsafe states ($U$ for $\str_\textrm{\it shield}^2$ or $\bar{S}_f$ for $\str_\textrm{\it shield}^1$), we select the action with the highest probability of returning to the safe set.

\subsubsection{Q-optimal Safety}
The final shield design we assess is based on dynamic programming, which results in the strongest guarantees for safety. We consider safety horizon $N \in \mathbb{N} \cup \{\infty \}$ and use a dynamic programming approach to compute the policy that minimizes probability of reaching unsafe set $\bar{S}_f$ in $N$ steps, resulting in optimal policy $\str^*_{S}$ and value function $V^*_{S}$, i.e.,
\begin{align*}
    & V^\str_{S}(s) = P(\omega_{\px}^N \in \Omega_\px^{\text{fin}, \str} \mid \omega_{\px}^N[0] \in s, \\
    & \hspace{38mm} \omega_{\px}^N [i] \in \bar{S}_f \text{ for some } i\in \mathbb{N}), 
\end{align*}
and $\str^*_S = \argmin_{\str} V^\str_S$ and $V^*_S = V^{\str_S^*}_S$.
When the shield is used, we assess the Q-value of each action and only allow actions that have a (unsafety probability) value below threshold $p$. Then, the shield is defined as
\begin{align*}
    \str_\textrm{\it shield}^\text{Q}(s) = \{a \in A \mid \sum_{s' \in \bar{S}} \bar{\Delta}(s,a,s') V^*_{S}(s') < p \}.
\end{align*}
In the event no action satisfies the bound $p$, the shield uses the optimal policy $\str_S^*$, i.e., if for a $s \in \bar{S}$, $\str_\textrm{\it shield}^\text{Q}(s) = \emptyset$, then $\str_\textrm{\it shield}^\text{Q}(s) = \str_S^*(s)$.
\begin{remark}
    During learning and deployment, we use post-posed shielding shown in Figure~\ref{fig:shield_arch}. Here the shield receives the observations of the environment and the action of the agent and returns a safe action to the environment. As in \cite{alshiekh2018safe}, there is no penalty (reward) associated with the shield changing the action.
\end{remark}

\section{Case Studies}
    \label{sec:experiments}

\begin{figure*}[t]
    \centering
    \begin{subfigure}[c]{0.43\textwidth}
        \includegraphics[width=\textwidth, trim={0 22mm 0 0}]{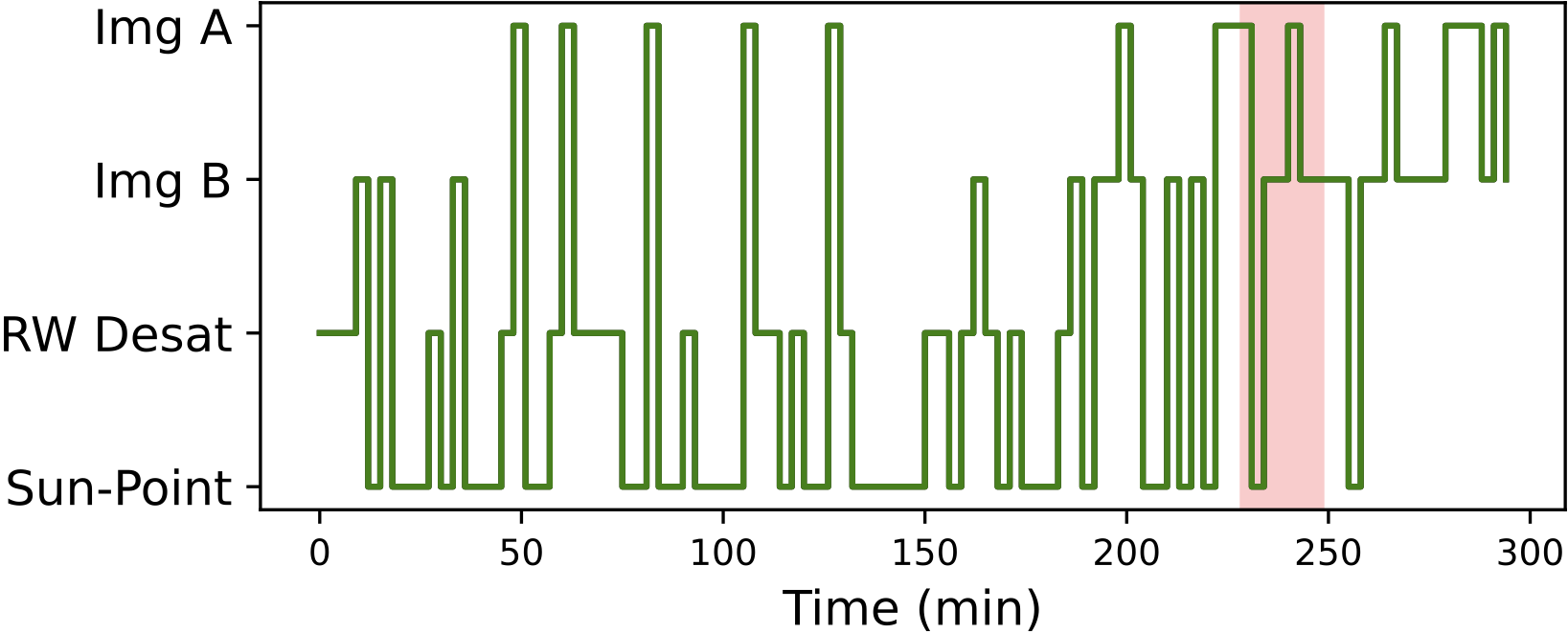}
    \end{subfigure}
    \qquad
    \begin{subfigure}[c]{0.43\textwidth}
        \includegraphics[width=\textwidth, trim={0 22mm 0 0}]{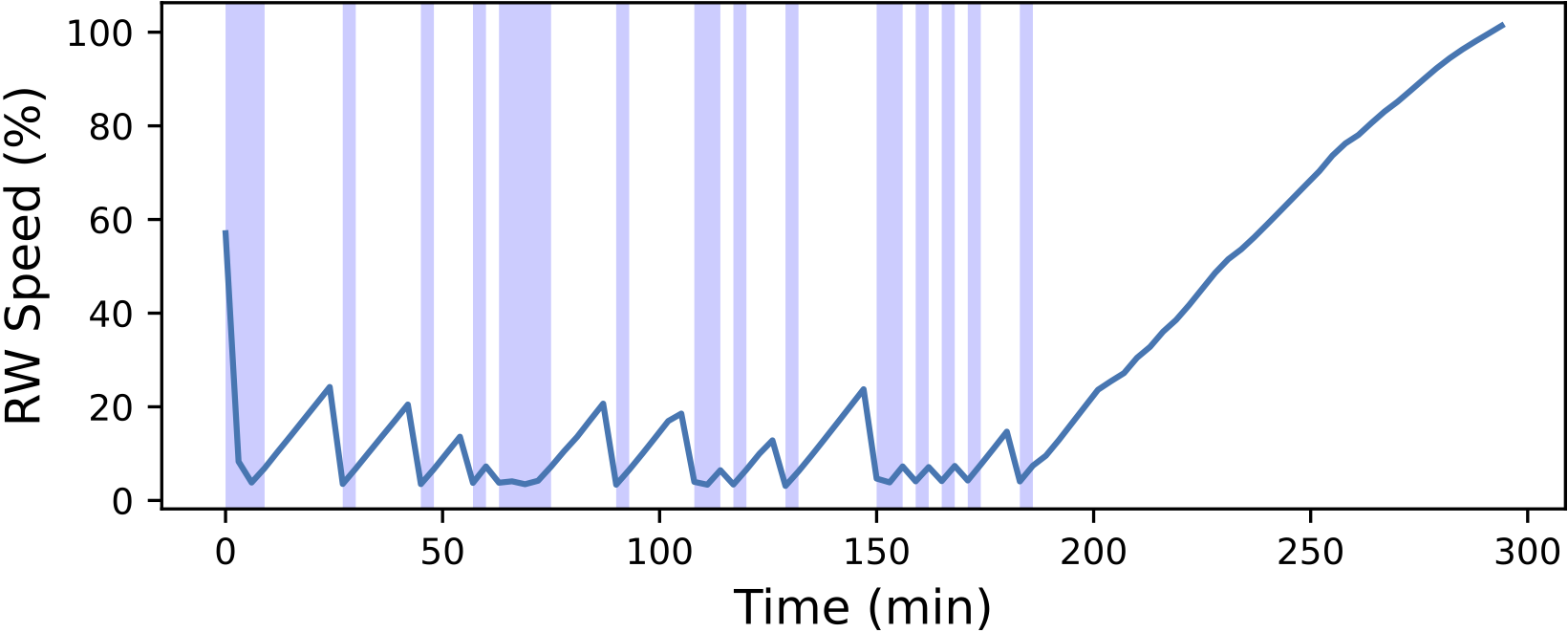}
    \end{subfigure}
    \begin{subfigure}[c]{0.43\textwidth}
        \includegraphics[width=\textwidth]{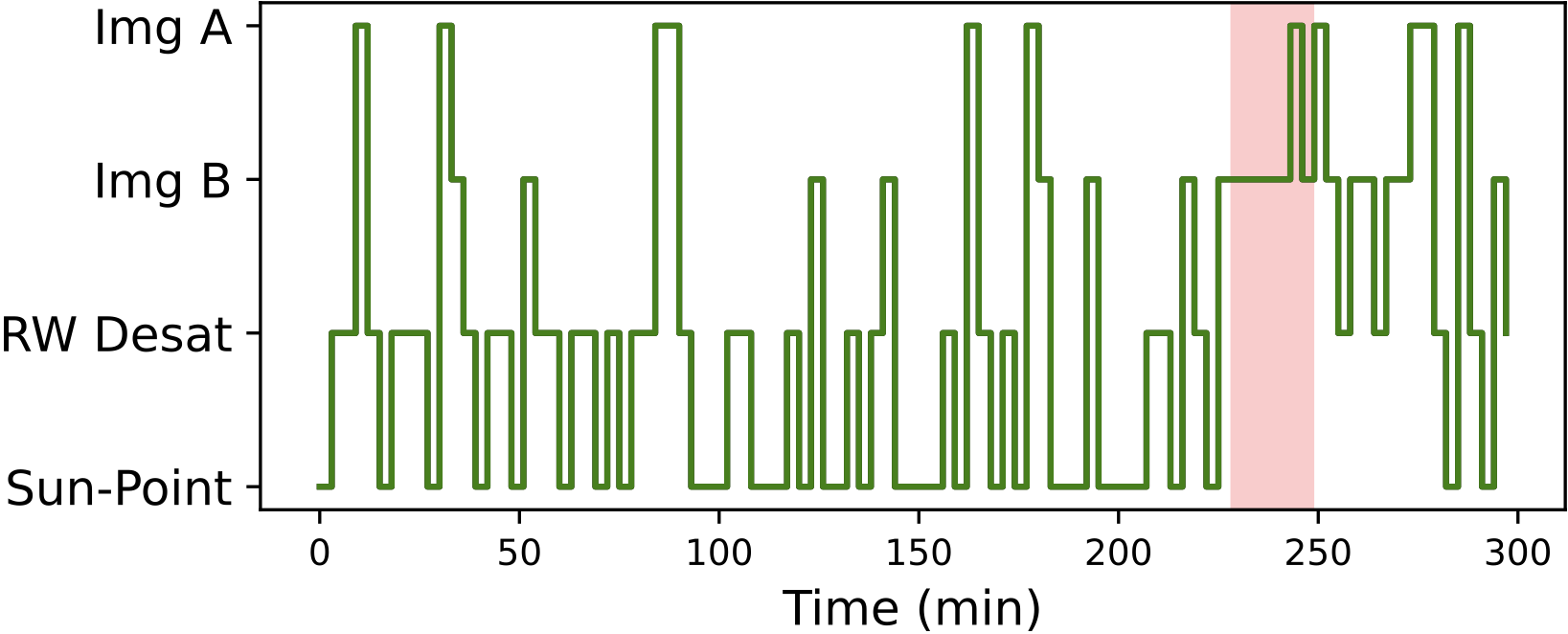}
    \end{subfigure}
    \qquad
    \begin{subfigure}[c]{0.43\textwidth}
        \includegraphics[width=\textwidth]{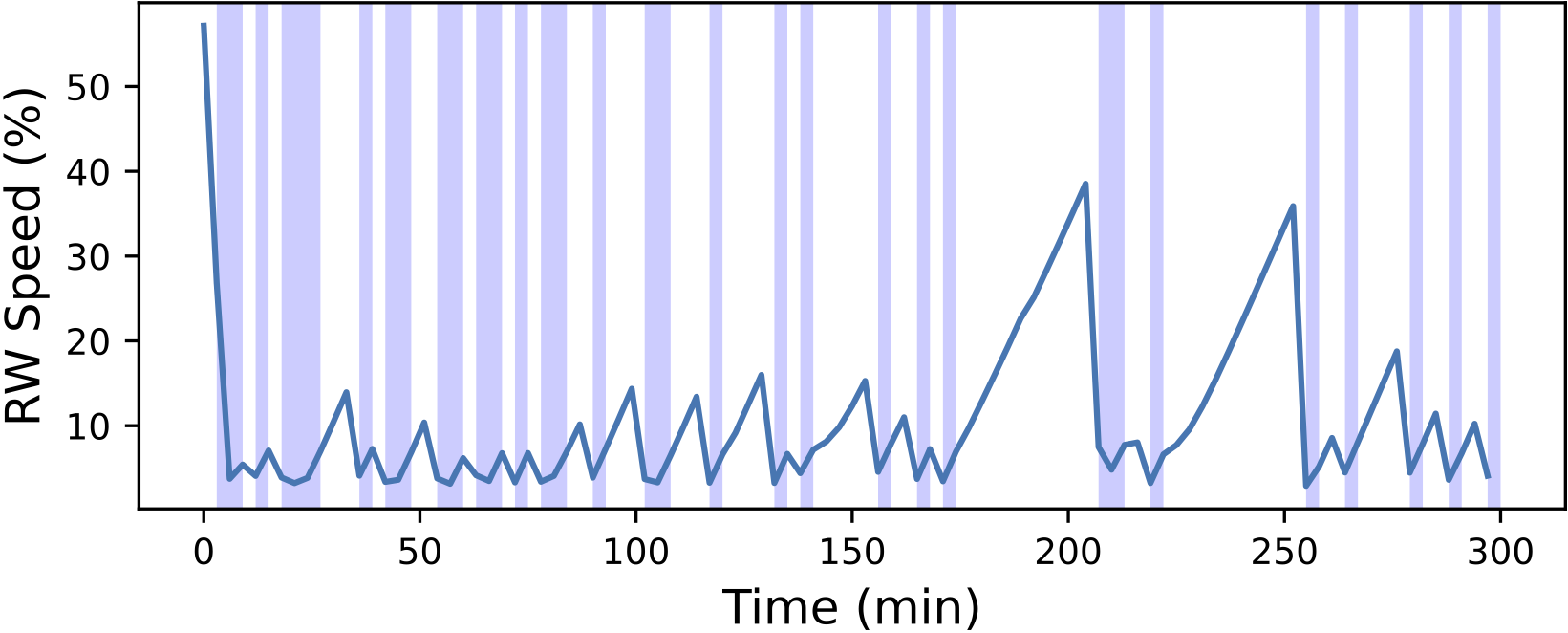}
    \end{subfigure}
    \caption{Action history and reaction wheel speeds when deploying under policy $\str_0$ (top) and $\str_1$ (bottom) from a fixed initial condition. The red highlight shows when the spacecraft has access to the target, the blue highlights show when the spacecraft is in Momentum Dumping (RW Desat) Mode. Note that policy $\str_1$ (trained in $\varphi_{0L} \land \varphi_S$) keeps the spacecraft safe after imaging the target whereas policy $\str_0$ (trained on only $\varphi_{0L}$) prioritizes imaging over spacecraft survival.}
    \label{fig:safety_spec_res}
    \vspace{-1mm}
\end{figure*}

We evaluate the efficacy of our LTL-SDRL framework on the spacecraft in Example~\ref{ex:sat}.
We first show the importance of training with safety specification on a simple task scenario.
Then, we consider a complex scenario for benchmarking (comparing) the three shield designs. The shields are each designed to prevent the spacecraft from having less than 20\% power and wheel speeds above 80\%. We also assess training with the shield and training without the shield.

In all the case studies, the Basilisk \cite{kenneally2020basilisk} simulator is used to create an environment for implementation of PPO2 \cite{schulman2017proximal}. The agent is trained with a learning rate $\alpha=3 \times 10^{-4}$ with a fully-connected network composed of two hidden layers of width 10 with a hyperbolic tangent activation function. The network is trained for $4.6 \times 10^5$ time steps on an Intel Core i7-12700K CPU at 3.60GHz with 32 GB of RAM limited to 8 threads. The input space of the network contains information on 19 states, composed as 13 states with 12 as in~\cite{harris2020spacecraft} and the addition of the DFA state. The Safety MDP consists of 100 discrete states.
All of our validations are computed from 1000 deployment simulations, where the dynamics are propagated with 0.5 second integration steps. Each flight mode is executed with and observations are received at three minute intervals. 

\subsection{Simple Task: Importance of $\varphi_S$ in Training}
We first demonstrate the effect of learning with and without a safety specification in (unshielded) DRL. We use the formula $\varphi_{0L}$ and $\varphi_S$ in \eqref{eq:simple_liveness_spec}-\eqref{eq:simple_safety_spec} from Example~\ref{ex:spec} and train for a fixed orbit and fixed target location. We compare the frequency of satisfying the specifications and the frequency of spacecraft failures over 300 minutes (100 time steps), which is longer than the time needed for one orbit (271 minutes). Here, we expect to see a high satisfaction rate of $\varphi_{0L}$ and some violations of $\varphi_S$. Results are shown in Table~\ref{table:simple_training}.

We first train on a DFA constructed from $\varphi_{0L}$ and refer to the policy returned from DRL as $\str_0$. The large average $V_F$ suggests frequent satisfaction of the specification during training under $\str_0$. When we deploy with $\str_0$, we see a high satisfaction rate of $\varphi_0$, but we also find frequent violations of $\varphi_S$ (i.e., either power falls below 20\% or reaction wheel speeds exceed 80\%) and even some spacecraft failures. This is expected, as without shielding and with no information in the reward structure about safety the policy never learns to avoid unsafe behavior.

\begin{table} [b]
    \caption{Training results for a simple task. Reported values are average value $V_F$ during training, rates of satisfaction of $\varphi_{0L}$ and violation of $\varphi_S$,  and spacecraft (SC) failure rate.}
    \label{table:simple_training}
    \begin{tabular} {c c c c c}
        \toprule
        Spec. & Avg. $V_F$ & \% Sat. $\varphi_{0L}$ & \% Violate $\varphi_{S}$ & SC Failure \\
        \midrule
        $\varphi_{0L}$ &  3.295 & 99.6 & 27.1 & 3.4 \\
        \midrule
        $\varphi_{0L} \wedge \varphi_S$ & 3.1 & 99.0 & 3.5 & 0 \\
        \bottomrule
    \end{tabular} 
\end{table}

When we train on a DFA constructed from $\varphi = \varphi_{0L} \wedge \varphi_S$, which implicitly incorporates the safety requirements into the reward structure, we find a policy $\str_1$. We see a similarly high satisfaction rate of $\varphi_{0L}$; however, very few runs violate $\varphi_S$ and no runs end with spacecraft failure. This demonstrates the power of incorporating safety into the training specification. We show a sample trajectory under both strategies in Figure~\ref{fig:safety_spec_res}, where each sample is initialized with the same parameters.

\subsection{Complex Tasks and Shielding}
We consider a more complex imaging task, to highlight the flexibility provided by training with LTL specifications and demonstrate the effects of shielding. Here we train to satisfy a task for a random LEO and random target locations. Our task is given as ``Image targets at least five times, alternating imaging modes starting from Imaging mode A. Only accept images that are taken when the attitude error ($|{\sigma}_{err}|$) is less than 0.008 radians and the attitude rate ($|\dot{\sigma}|$) is less than 0.002 radians per second." This is translated into the co-safe LTL formula
$$\varphi_{1L} = \F(\prop_3 \wedge \X \F(\prop_4 \wedge \X \F(\prop_3 \wedge \X \F(\prop_4 \wedge \X \F \prop_3))))$$
where $\prop_3$ and $\prop_4$ are similar to $\prop_0$ from Example~\ref{ex:spec}, but the corresponding regions only include one imaging mode. Note that in traditional reward design a finite state machine would need to be constructed to account for mode switching; however, in our formulation the switching is automatically captured through the specification.
The safety specification is the same $\varphi_S$ in \eqref{eq:simple_safety_spec} as above, and probability threshold is $p = 0.05$.

For this scenario, each episode consists of 90 actions which corresponds to three orbits. We assess the average value $V_F$ during training, the satisfaction rate of specification with and without the shield after the agent has been trained, the average number of shield interventions, and the frequency of unsafe behavior. Results are shown in Table~\ref{table:training_res}.

\begin{table*}[t]
    \centering
    \caption{Results for the complex specification $\varphi_{1L}$. We report average value $V_F$ of trajectories during training, liveness specification satisfaction rate (\% Sat. $\varphi_{1L}$), safety specification violation rate (\% Violate $\varphi_S$), spacecraft failure rate (SC Failure, \% of tests), and the average number of shield interventions (over 90 time steps) when $\varphi_{1L}$ was satisfied and not satisfied. $^{*1}$ and $^{*2}$ denote that the same policy was used for these experiments, i.e., they were trained without a shield. 
    }
    \label{table:training_res}

    \begin{tabular} { l  c  c  l  c  c  c  c c }
        \toprule
        \multirow{2}{*}{Shield Type} & Trained & \multirow{2}{*}{Spec.} & \multirow{2}{*}{Avg. $V_F$} & \% Sat. & 
        \% Violate & \multirow{2}{*}{SC Failure} &
        \multicolumn{2}{c}{\underline{\quad Avg. \# of Interventions \quad}}\\
         & w/ Shield & & & $\varphi_{1L}$ &  $\varphi_S$ & & Sat. $\varphi_{1L}$  & Not Sat. $\varphi_{1L}$\\
        \midrule
        \multirow{2}{*}{No Shield} & No$^{*1}$ & $\varphi_{1L}$ & \; 3.39$^{*1}$  & 92.7 & 55.3 & 10.9 & -- & --  \\
                                   & No$^{*2}$ & $\varphi_{1L} \wedge \varphi_S$ & \; 2.47$^{*2}$ & 90.9 & 11.7 & 1.4 & -- & -- \\
        \midrule
        \multirow{4}{*}{One Step ($\str_\textrm{\it shield}^1$)} & No$^{*1}$ & $\varphi_{1L}$ & \; 3.39$^{*1}$ & 79.2 & 0.4 & 0 & 13.6 & 29.2 \\
                                                      & No$^{*2}$ & $\varphi_{1L} \wedge \varphi_S$ & \; 2.47$^{*2}$ & \textbf{85.9} & 0.6 & \textbf{0} & \, \textbf{3.7} & \textbf{11.1} \\
                                                      & Yes \;\, & $\varphi_{1L}$ & \; 1.88 & 64.8 & 0.8 & 0 & 44.0 & 60.6 \\
                                                      & Yes \;\, & $\varphi_{1L} \wedge \varphi_S$ & \; 1.93 & 74.9 & 0.6 & 0 & 48.6 & 66.8 \\
        \midrule
        \multirow{4}{*}{Two Step ($\str_\textrm{\it shield}^2$)} & No$^{*1}$ & $\varphi_{1L}$ & \; 3.39$^{*1}$ & 82.0 & 0.6 & 0 & 13.2 & 28.5 \\
                                                      & No$^{*2}$ & $\varphi_{1L} \wedge \varphi_S$ & \; 2.47$^{*2}$ & \textbf{86.2} & 0.9 & \textbf{0} & \, \textbf{3.9} & \textbf{10.9} \\
                                                      & Yes \;\, & $\varphi_{1L}$ & \; 1.84 & 61.6 & 1.0 & 0 & 41.4 & 59.9 \\
                                                      & Yes \;\, & $\varphi_{1L} \wedge \varphi_S$ & \; 1.86 & 72.6 & 1.2 & 0 & 47.1 & 64.0 \\
        \midrule
        \multirow{4}{*}{Q-optimal ($\str_\textrm{\it shield}^\text{Q}$)} & No$^{*1}$ & $\varphi_{1L}$ & \; 3.39$^{*1}$ & 79.9 & 0.2 & 0 & 14.2 & 28.9 \\
                                                             & No$^{*2}$ & $\varphi_{1L} \wedge \varphi_S$ & \; 2.47$^{*2}$ & \textbf{85.2} & 0.4 & \textbf{0} & \, \textbf{3.8} & \textbf{11.2}  \\
                                                             & Yes \;\, & $\varphi_{1L}$ & \; 1.70 &  63.2 & 0.5 & 0 & 44.2 & 61.0\\
                                                             & Yes \;\, & $\varphi_{1L} \wedge \varphi_S$ & \; 2.23 & 73.4 & 0.2 & 0 & 49.1 & 66.9 \\
        \bottomrule
    \end{tabular}
    \vspace{-12pt}
\end{table*}

\subsubsection{Trained and Deployed without Shielding} We see a similar trend as with the simple specification, where training with the safety specification included results in a lower reward, fewer spacecraft failures, and fewer safety violations. Due to the more stressing environment, we see more safety violations during training which results in the large gap in value $V_F$ between the two policies. As noted earlier, we still see a correlation between the average value $V_F$ of trajectories and the probability of satisfying the liveness specification, a higher $V_F$ means the policy is more likely to result in satisfaction when deployed.

\subsubsection{Trained without Shielding and Deployed with Shielding}
We first note that in each case a shield is deployed, there are no instances of spacecraft failure and the frequency of violations of $\varphi_S$ fall below 5\%, which is the threshold we allow for choosing safe actions. When we train with just $\varphi_{1L}$, we see a significant reduction in satisfaction rate and a high frequency of shield interventions. This follows intuition, as over 50\% of trajectories would violate $\varphi_S$ without shielding. Contrarily, when we train on $\varphi_{1L} \wedge \varphi_S$, we see far fewer shield interventions and a higher rate of satisfaction of $\varphi_{1L}$. This demonstrates that the policy learned from the safety specification acts in a more flexible manner when shielded. In all cases, we see that far more shield interventions occur when a trajectory does not satisfy $\varphi_{1L}$ as actions must be repeatedly corrected to maintain guarantees.

\subsubsection{Trained and Deployed with Shielding} Minor differences are seen between our shield designs when we train with them in the loop. As noted in~\cite{alshiekh2018safe}, the policy learns to rely on the shield interventions. In each case, over half of all actions the DRL policy chooses are rejected by the shield in place of a safe action which results in far fewer trajectories satisfying $\varphi_{1L}$. 
Here, the improvements seen when training with a safety specification are more noticeable, despite the same number of shield interventions occurring. This, again, shows the correlation between $V_F$ and the satisfaction rate.

\begin{remark}
Despite the differences in guarantees provided by each shield, we see that they perform very similarly when deployed. This suggests that the safety MDP we constructed contains overly conservative transition probabilities, resulting in similar policies among the shields. The states chosen to represent safety likely increase conservatism, as these states have highly non-deterministic transitions which requires the shield to select actions for worst case orientations (e.g. The norm of the wheel speed vector contains no information about direction). 
\end{remark}

\section{Conclusion}
    \label{sec:conclusion}

In this work, we provide a method to enable autonomous decision making for the satisfaction of complex Earth imaging tasks through SDRL. Safety is ensured through the designs of three shield algorithms, which are based on a simulated safety MDP. We identify that training on composed liveness and safety specifications without a shield results in the best interaction of policy and shield, i.e., a high rate of satisfaction of liveness, few violations of safety, and few shield interventions. The shields we produced are far less restrictive than prior designs.  However, while safety guarantees hold for the safety MDP, the transition probabilities are computed based on empirical evaluation (physics simulator). Numerical simulations suggest the shields provide stronger guarantees than necessary. We hope to further formalize the safety MDP in future work and reduce the associated conservatism seen by the shielding.

\bibliographystyle{IEEEtran}
\bibliography{references}

\end{document}